# Evaluating Empathetic Chatbots in Customer Service Settings

**Akshay Agarwal[1], Shashank Maiya[2], Sonu Aggarwal[3]**
[1] KMK Consulting, Inc.
[2] ServiceNow, Inc.
[3] Unify Square, Inc.

aa2657@cornell.edu, shashank.maiya@gmail.com, sonuag@unifysquare.com

## Abstract

Customer service is a setting that calls for empathy in live human agent responses. Recent advances have demonstrated how open-domain chatbots can be trained to demonstrate empathy when responding to live human utterances. We show that a blended skills chatbot model that responds to customer queries is more likely to resemble actual human agent response if it is trained to recognize emotion and exhibit appropriate empathy, than a model without such training. For our analysis, we leverage a Twitter customer service dataset containing several million customer<->agent dialog examples in customer service contexts from 20 well-known brands.

## 1 Introduction

The area of open-domain chatbots has emerged as an exciting area of NLP research in recent years. On the one hand, open-domain SOTA chatbot models can generate very impressive examples of human-model conversations, capable of being very engaging, conveying a rich persona, holding their own in conversations, and even demonstrating some reasoning and some ability to generate humor. On the other hand, open-domain chatbots are still nowhere close to being human-like in some dimensions; they tend to be prone to repetition, lean towards simple responses, and lack any deep understanding of the subject matter.

The central hypothesis of this paper is that a language model finetuned on emotions (per the Empathetic Dialogues (ED) work in Rashkin et al 2018) has better performance on customer service task-oriented queries than one without such finetuning. The code for our work can be found on Github[1]

## 2 Related Work

SOTA chatbot models tend to leverage transformers (Vaswani et al., 2017) and poly-encoder variations of transformers (Humeau et al., 2019).

Blended skill chatbots have been shown to have superior performance over single-skill chatbots in open-domain tasks. Such skills can include inference, personalization, empathy, and knowledge. (Welleck et al., 2018) develop the Dialogue Natural Language Inference (DialogueNLI) technique to improve the consistency of dialogue generation. Models can be trained to exhibit a consistent persona, a set of personality traits consistently referenced in their responses to make their dialogues with actual humans more engaging (Zhang et al., 2018). Similarly, models can be trained to demonstrate empathy and talk about emotional personal situations, as in the work on Empathetic Dialogues (Rashkin et al., 2019). Models can be made to be knowledgeable and discuss a topic in depth, as in the Wizard of Wikipedia task (Dinan et al., 2019). Such disparate skills can be effectively combined to obtain a model that blends all skills into a single conversational agent, as shown in the BlendedSkillTalk work of Smith et al., 2020. Such blended skill chatbots have been shown to be SOTA on several measures, generating impressive

---
[1] https://github.com/shashankvmaiya/Task-Oriented-Chatbot-With-Empathy

Table 1: Twitter Customer Care Dataset Examples

examples of open-domain conversations with humans (Roller et al., 2020).

Evaluating such models can be a challenge, due to the large variability in "soft skills" that make these models particularly suitable for human evaluation (Li et al., 2019). However, Perplexity has emerged as a leading computable metric that has been shown to be a reasonable approximation of human evaluation of the quality of chatbot responses (Adiwardana et al., 2020).

In our work, we evaluate performance of transformer models trained on empathy, leveraging the Empathetic Dialogues work of Rashkin et al., 2019, against actual customer service exchanges between customers and human agents. We evaluate such models using Perplexity, along the lines of (Adiwardana et al., 2020). We generally use the modeling techniques of (Roller et al., 2020), but evaluate our models based on the automated techniques of (Adiwardana et al, 2020).

## 3 Data

We use the Twitter Customer Care dataset, available on Kaggle, as the foundation of our work. As described by its authors, this dataset is "a large, modern corpus of tweets and replies to aid innovation in natural language understanding and conversational models, and for the study of modern customer support practices and impact". This dataset consists of over 2.8 million user tweets and customer service responses from 20 well-known brands that leverage Twitter for customer support, including Delta Air Lines, Apple, British Airways,

---

Original Text: `Are the rumors on the OnePlus 5t true? I was looking at getting the 5 but if a new version is coming I might as well wait @O2 @123612`

Original Label: `@123611 Hey there, no news at the moment but register here to be notified when there is: https://t.co/UcPgezkL14  üëç`

Preprocessed Example compliant with ParlAI framework:

`text:are the rumors on the oneplus 5t true ? i was looking at getting the 5 but if a new version is coming i might as well wait    labels:hey there , no news at the moment but register here to be notified when there is :    episode_done:True`

Figure 1: Preprocessed example from Twitter Customer Care dataset

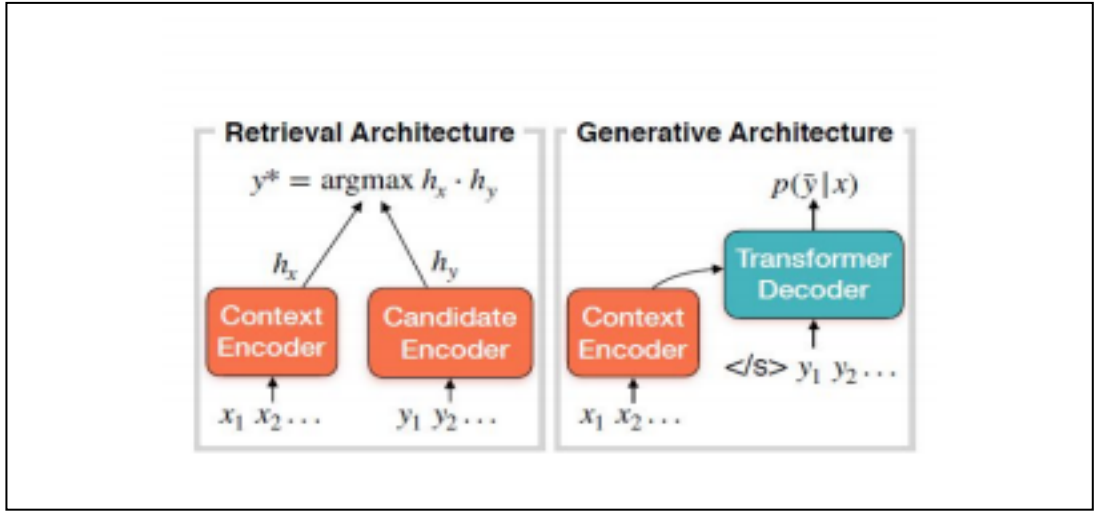

Figure 2: Retrieval and Generative Architecture

Chipotle, Amazon, UPS, etc., thereby spanning a number of industry verticals.

While we will primarily use this dataset for baselining and evaluation, we will use an additional dataset - Empathetic Dialogues (or ED) - for fine-tuning models for emotion detection. The dataset and approach are both described in (Rashkin et al., 2018). This ED dataset consists of 25k chats classified by underlying emotion.

Additionally, we will use models pre-trained on the pushshift.io Reddit dataset, consisting of 1.7B Reddit chats - an often-referenced dataset in the literature. We will not directly interact with this dataset but use models already pre-trained on it.

The Twitter dataset is in CSV format (a snapshot shown in Table 1), where each row is a tweet. We will preprocess the data to remove hashtags and other punctuation. Other preprocessing steps will include eliminating non-English exchanges and mapping some Twitter "slang" terms (e.g. commonly used abbreviations such as "LOL") to actual words in our vocabulary. An example of the original text from customer and the response from customer care along with the preprocessed conversation exchange which is compliant with the ParlAI format is shown in Figure 1

We will split the 2.8M-record Twitter Customer Care Dataset into train/valid/test sets (initially aiming for a 60/10/30 split) and use the test set to evaluate our results.

## 4 Models

The two most common approaches for open-domain chatbots are generative approaches and retrieval approaches. Generative models predict the next utterance similar to a language model, conditioning on the dialogue history and generating a response one word at a time. Retrieval, or ranking, models have the model choose the best response from a large corpus of actual responses (e.g. the pushshift.io Reddit dataset has O(1B) responses), determining "best response" by encoding both the output of a generative step (BERT or transformer) and the corpus of responses, computing similarity to identify the closest response. Figure 2 illustrates these architectures.

### 4.1 Generative Transformer

We use the transformer generator architecture for fine-tuning. It is a generative-based agent that performs seq2seq encoding/decoding with transformer encoders/decoders. The transformer decoder uses the encoder output to predict a sequence of words y and is trained to minimize the negative log-likelihood of the target sequence. At inference time, we use beam search with a beam size of 3 for next utterance prediction.

| Model Type | Architecture Type | Dataset | Parameters | Embedding Size | Heads | Layers |
|---|---|---|---|---|---|---|
| Generative | Transformer Generator | Reddit (1.7B examples) | 87 M | 512 | 16 | 8 |
| Retrieval | Transformer bi-encoder | Reddit (174M examples) | 256 M | 768 | 12 | 12 |
| Retrieval | Transformer poly-encoder | Reddit (174M examples) | 256 M | 768 | 12 | 12 |

Table 2: Parameters pertaining to the Pretrained Baseline Models

We used the Pretrained Transformer model available in the zoo library in the ParlAI as our base model for fine-tuning. This was trained on 1.7 billion Reddit conversations. The model is based on the Transformer architecture (Vaswani et al., 2017) and contains 87 M parameters. Pretrained models without any fine-tuning on Twitter Datasets and Empathy datasets will be referred to as "*pretrained_baseline*" hereafter. The transformer networks used in most experiments have the same base architecture - 8 layers and 16 transformer heads. The embedding dimensions were set to 512 and the input text length was truncated to 400. A batch size of 24 was used to train the data and evaluate the results. This model will be our baseline evaluation. The details of this pretrained model is provided in Table 2.

We created three models based on a generative transformer architecture to evaluate our results. All three models leverage the existing pretrained model (1.7 billion Reddit conversations). The first model is fine-tuned on the Empathetic Dialogues (ED) dataset, referred to hereafter as *finetuned_ed*. The second model is fine-tuned on the Twitter customer care dataset, hereafter referred to as *finetuned_cc*. We have also fine-tuned the *finetuned_ed* model on the customer care dataset, this being hereafter referred to as the *finetuned_ed_cc* model. All generator models employ BPE encoding (Sennrich et al., 2016) which is effective at enabling generators to copy rare words.

We use a history size of three for fine-tuning our models – i.e. in an episode, we take into account the previous three occurrences (wherever possible) as our input and the corresponding label as our output.

## 4.2 Retrieval Transformer

In the retrieval-based architecture, the model is given a large set Y of candidate responses and picks the "best" one. A retrieval-based agent that encodes a context sequence and a candidate sequence with separate BERT-based (Devlin et al. (2019)) Transformers. The candidate is chosen via the highest dot-product score between the context and candidate encodings.

We use two baseline architectures defined in Humeau et al (2020) to evaluate retrieval based models – a.) transformer biencoder and b.) transformer polyencoder. These models are also trained on reddit examples (model details are provided in Table 2). The pretrained model has the same architecture as BERT-base consisting of 12 layers, 12 attention heads, and a hidden size of 768. The input text was truncated to 360. We used the adamax optimizer with a batch size of 10. Similar to generative-based architecture we built three models to evaluate our results. The first model is fine-tuned on Empathetic Dialogues (ED) dataset (*finetuned_ed*), the second model fine-tuned on Customer Care (CC) dataset (*finetuned_cc*), and the final model is fine-tuned on both the ED and CC datasets (*finetuned_ed_cc*).

## 5 Experimental Evaluation

We use the Parl.AI framework of Miller et al 2017, an open-source software platform by Facebook for dialog research implemented in Python. It provides a unified framework for training and testing dialog models, evaluation, and a repository of machine learning models for comparing with other models.

| Model Architecture | Model | PPL | Hits@1 |
|---|---|---|---|
| **RETRIEVAL: Transformer/biencoder** | pretrained_baseline | | 29 |
| | finetuned_ed | | 41 |
| | finetuned_cc | | 56.94 |
| | finetuned_ed_cc | | **57.18** |
| **RETRIEVAL: Transformer/polyencoder** | pretrained_baseline | | 20 |
| | finetuned_ed | | 35 |
| | finetuned_cc | | 52.21 |
| | finetuned_ed_cc | | **54.53** |
| **GENERATIVE: Transformer/generator** | pretrained_baseline | 28.64 | |
| | finetuned_ed | 33.06 | |
| | finetuned_cc | **10.01** | |
| | finetuned_ed_cc | **10.01** | |

Table 3: Results. Automatic evaluation metrics on the test set. Pretrained_baseline: model pretrained on a dump of 1.7 billion REDDIT conversations. Finetuned_ed, finetuned_cc, finetuned_ed_ecc: models fine-tuned over the EmpatheticDialogues training data (ED), CustomerCare training data (CC), or both (ED+CC). PPL: Perplexity. Hits@1: Precision retrieving the correct test candidate out of 20 test candidates.

The ED dataset is available within the ParlAI framework, which we have used extensively for our work. It provides an integrated environment to evaluate and fine-tune chat models against an extensive set of included chat datasets referenced in our Literature Review. We imported the Twitter dataset into this ParlAI framework, preprocessing several elements – lowercasing, removing emojis, removing emoticons, removing URLs & HTML tags, converting chat word acronyms to full words, removing twitter handles and related artifacts, separating out punctuation with spaces, removing all non-English sentences, and finally converting each dialogue "episode" into the format consumable by ParlAI – an example of the preprocessing is shown in Figure 1.

### 5.1 Evaluating Generative Models

We evaluated 4 model/dataset combinations and all the models were evaluated on TWCS (Twitter Customer Service) test dataset.

1. Pretrained_baseline: Transformer/generator pretrained on Pushshift.io Reddit
2. Finetuned_ed: The model of #1 fine-tuned on Empathetic Dialogues (available in ParlAI through Rashkin et al., 2018)
3. Finetuned_cc: The model of #1 fine-tuned on TWCS (train)
4. Finetuned_ed_cc: The model of #2 fine-tuned on TWCS (train)

We evaluate our generative models on perplexity (PPL) of the gold response and were able to reduce it to 10 from 54. Our results are summarized in Table 3.

### 5.2 Evaluating Retrieval Models

We used two model baseline architectures to evaluate retrieval models – 1.) Biencoder Transformer and 2.) Polyencoder Transformer. For each of the baseline model, we fine-tune based on ED and CC dataset to obtain {Finetuned_ed, Finetuned_cc, Finetuned_ed_cc} models and test on TWCS test dataset.

Batchsize of 10 was chosen and all labels in a batch are chosen as candidates to train the model, i.e., for a batch size of 10 the model has to choose the correct response from 9 distractors and one gold response. We evaluated our model on Hits@1 and the results can be found in Table 3

| EXAMPLE 1 | | |
|---|---|---|
| **Customer Text** | **Models** | **Model Response** |
| extremely disappointed with comcast . i was told i could get xfinity . tech came twice ( two personal days from work ) and no cable was installed outside beforehand . then survey done for a cable to be put in from road to house . | Pretrained_baseline | i ' ve had comcast for a few years now and have never had a problem with them . |
| | Finetuned_ed | that ' s terrible . i ' m sorry to hear that . |
| | Finetuned_cc | i ' m sorry to hear about the service issues . please dm your account number and phone number so i can look into this for you . |
| | Finetuned_ed_cc | i ' m sorry to hear about the service issues . i would be happy to look into this for you . please dm your account number and phone number . |
| | Gold Response | i want to address your concerns . did you need for us to schedule someone to come out to install those lines ? |
| EXAMPLE 2 | | |
| **Customer Text** | **Models** | **Model Response** |
| trying to reach a representative about my recently deceased fathers flight ; i need customer relations but there's no one there to help ! | Pretrained_baseline | i ' m sorry to hear that . i hope you find someone to help you . |
| | Finetuned_ed | oh no ! i ' m sorry to hear that . i hope you can get through it . |
| | Finetuned_cc | we ' re here to help ! send us a dm with your email address so we can connect . |
| | Finetuned_ed_cc | we ' re sorry to hear about the loss of your father . we ' d like to help you with this . please dm us your contact info and we ' ll get started . |
| | Gold Response | thanks for reaching out , dylan . our customer relations team has not yet resumed normal operations due to the holidays . what seems to be going on ? |
| EXAMPLE 3 | | |
| **Customer Text** | **Models** | **Model Response** |
| any chance of knowing why my train was cancelled ? | Pretrained_baseline | i don ' t know why it was cancelled , but i ' m pretty sure it ' s because it was delayed . |
| | Finetuned_ed | i don ' t know . i ' m not sure what happened . |
| | Finetuned_cc | hi , sorry to hear this . please can you dm your confirmation number ? |
| | Finetuned_ed_cc | hi there , sorry to hear this . what service are you on ? |
| | Gold Response | hi mike , what service were you looking to get ? |

Table 4: Comparing different model responses for transformer/generative architecture

| Model Architecture | Model | Train Time (hours) | Epochs Trained |
|---|---|---|---|
| RETRIEVAL: Transformer/biencoder | pretrained_baseline | | |
| | finetuned_ed | 2.68 | 5 |
| | finetuned_cc | 4.66 | 5.5 |
| | finetuned_ed_cc | 4.63 | 5.5 |
| RETRIEVAL: Transformer/polyencoder | pretrained_baseline | | |
| | finetuned_ed | 5.01 | 9 |
| | finetuned_cc | 3.81 | 4.5 |
| | finetuned_ed_cc | 4.23 | 5 |
| GENERATIVE: Transformer/ generator | pretrained_baseline | | |
| | finetuned_ed | 3.72 | 3.5 |
| | finetuned_cc | 7.97 | 7 |
| | finetuned_ed_cc | 7.94 | 7 |

Table 5: Training resources consumed for different models

## 6 Results

Our results consisting of perplexity for generative models and hits@1 for retrieval models are summarized in Table 3. We trained our model on a TeslaK80 GPU with 12 GB memory. Details about training resources is summarized in Table 5.

### 6.1 Generative Models

Generative models did not appear to benefit from fine-tuning on ED: the perplexity for both the finetuned_cc and finetuned_ed_cc versions were 10.01.

### 6.2 Retrieval Models

Evaluation results on retrieval models bear out our hypothesis: models fine-tuned on the ED and CC datasets perform better (measured by Hits@1) than models fine-tuned on only CC. The improvements obtained for Transformer/biencoder models were relatively modest, at 0.24% improvement. However, Transformer/polyencoder models appeared to show significant improvement, increasing Hits@1 by 4.44%, from 52.21 to 54.53.

## 7 Analysis

For retrieval models specifically (both the biencoder and polyencoder versions), the model pre-fine-tuned on Empathetic Dialogues (Rashkin et al. 2018) and subsequently fine-tuned on the Twitter Customer Service Dataset (train) performed better than the model that was not trained on Empathetic Dialogues, thus bearing our hypothesis. This demonstrates that chatbot trained to recognize and exhibit empathy appears to more closely mimic actual customer service agent responses in response to actual customer service queries.

However, for the Generative models, there was no apparent benefit in pre-training on Empathetic Dialogues. The reasons are not readily apparent.

A qualitative comparison of the outputs of the different models using generative transformer architecture (automated customer care responses) in response to the original customer queries (Table 4) is illuminating.

- The pretrained transformer models have the simplest responses, usually not directly addressing the customer query (and, in some cases, opposing the customer's sentiment).

- The ED-trained models consistently show high empathy; however, it still doesn't really try to solve the issue that the customer raises.
- The CC-trained models consistently appear to focus "on task", i.e. addressing the customer need. It tends to ask questions or requests additional information aiming at solving the issue that the customer is facing.
- The ED-CC trained models try to combine empathy (which often is a sorry statement at the beginning) with task focus.

Further work can extend this analysis to assess whether empathetic chatbots fare better in task-oriented situations, when evaluated against typical task-oriented metrics (e.g. task completion rates).

# 8 Conclusion

In this work, we present experimental evidence that supports the hypothesis that chatbots trained to recognize and express emotions/empathy more closely resemble actual human agent responses to actual customer service queries, relative to chatbots without such training.

**Acknowledgments**

We would like to thank Prof. Christopher Potts, our Course Facilitator Powell Molletti, and the staff of XCS224U for their guidance throughout this project.

**Authorship Statement**

Akshay Agarwal – identifying Twitter dataset, experiment structure, model evaluation code for key scenarios.

Shashank Maiya – methodology review, initial environment, main code structure and Parl.Ai framework code

Sonu Aggarwal – end-to-end coordination, authored drafts of paper and of background assignment papers